\documentclass[sigconf,letterpaper,nonacm]{acmart}
\AtBeginDocument{%
  }

\setcopyright{acmlicensed}
\copyrightyear{2025}
\acmYear{2025}
\setcopyright{acmlicensed}\acmConference[GeoAnomalies '25]{The 2nd ACM SIGSPATIAL International Workshop on Geospatial Anomaly Detection}{November 3--6, 2025}{Minneapolis, MN, USA}
\acmBooktitle{The 2nd ACM SIGSPATIAL International Workshop on Geospatial Anomaly Detection (GeoAnomalies '25), November 3--6, 2025, Minneapolis, MN, USA}
\acmDOI{10.1145/3764914.3770888}
\acmISBN{979-8-4007-2260-8/2025/11}

\usepackage[T1]{fontenc}
\usepackage{xspace}
\newcommand{\systemname}{\systemnamens\xspace}
\newcommand{\systemnamens}{BeSTAD}

\begin{document}

\title[\systemname]{\systemname: Behavior-Aware Spatio-Temporal Anomaly Detection for Human Mobility Data}

\author{Junyi Xie, Jina Kim, Yao-Yi Chiang}
\email{{xie00422, kim01479, yaoyi}@umn.edu}
\affiliation{%
  \institution{University of Minnesota}
  \city{Minneapolis}
  \state{Minnesota}
  \country{USA}
}

\author{Lingyi Zhao, Khurram Shafique}
\email{{lzhao, kshafique}@novateur.ai} 
\affiliation{%
  \institution{Novateur Research Solutions}
  \city{Ashburn}
  \state{Virginia}
  \country{USA}}

\renewcommand{\shortauthors}{Xie et al.}

\begin{abstract}
Traditional anomaly detection in human mobility has primarily focused on trajectory-level analysis, identifying statistical outliers or spatiotemporal inconsistencies across aggregated movement traces. However, detecting individual-level anomalies, i.e., unusual deviations in a person's mobility behavior relative to their own historical patterns, within datasets encompassing large populations remains a significant challenge. In this paper, we present \textbf{\systemname} (\textbf{Be}havior-aware \textbf{S}patio-\textbf{T}emporal \textbf{A}nomaly \textbf{D}etection for Human Mobility Data), an unsupervised framework that captures individualized behavioral signatures across large populations and uncovers fine-grained anomalies by jointly modeling spatial context and temporal dynamics. \systemname learns semantically enriched mobility representations that integrate location meaning and temporal patterns, enabling the detection of subtle deviations in individual movement behavior. \systemname further employs a behavior-cluster-aware modeling mechanism that builds personalized behavioral profiles from normal activity and identifies anomalies through cross-period behavioral comparison with consistent semantic alignment. Building on prior work in mobility behavior clustering, this approach enables not only the detection of behavioral shifts and deviations from established routines but also the identification of individuals exhibiting such changes within large-scale mobility datasets. By learning individual behaviors directly from unlabeled data, \systemname advances anomaly detection toward personalized and interpretable mobility analysis.
\end{abstract}




\maketitle

\vspace{-.1in}

\section{Introduction}

Understanding anomalies in human mobility patterns is vital across domains such as epidemiology, transportation, and urban planning, where irregular movement behaviors can reveal meaningful deviations in social, environmental, or operational dynamics~\cite{liu2020online, zhang2011ibat}. Traditional approaches to trajectory anomaly detection primarily focus on identifying statistical outliers or spatiotemporal inconsistencies within individual trajectories or stay points~\cite{liu2020online, zhang2011ibat}. For instance, a vehicle taking an unexpected detour or a pedestrian deviating from a known route might be flagged as anomalous. Such methods effectively capture \textit{where} and \textit{when} a specific trip diverges from expected paths, but they treat each trajectory as an independent instance and overlook how an \textit{individual's overall behavior} may change across time ~\cite{wen2024uncertainty, hu2024context, duan2024back}. In many real-world settings, however, the more critical question is which individuals exhibit consistent or contextually meaningful shifts in their long-term mobility, such as a commuter whose daily routes gradually move toward a new region, or a delivery worker whose patterns abruptly change due to altered assignments or lifestyle shifts.  

Despite recent progress in trajectory anomaly detection, existing methods face two key limitations when extended to the individual level. First, multi-scale spatial semantics remain underexploited. Most frameworks rely on trajectory reconstruction or prediction accuracy and fail to fully incorporate the rich contextual meaning of  locations~\cite{liu2020online, ying2011semantic}. Consequently, they struggle to recognize \textit{location-semantic anomalies}, i.e., behaviors inconsistent with the functional semantics of visited places, such as a typical office worker frequently visiting industrial facilities during working hours. Second, individualized behavior modeling remains insufficient. Although recent studies such as Hu~\citet{hu2024context} incorporate individual identifiers into model training, they often overfit to aggregate population patterns, missing subtle but meaningful personal deviations. For example, a night shift healthcare worker commuting at midnight might be incorrectly flagged as anomalous because such activity diverges from the dominant population trend. 

To address these challenges, we propose \textbf{\systemname} (\textbf{Be}havior-aware \textbf{S}patio-\textbf{T}emporal \textbf{A}nomaly \textbf{D}etection for Human Mobility Data), an unsupervised framework for fine-grained individual-level anomaly detection in large-scale human mobility data. \systemname jointly models spatial semantics and temporal dynamics to learn how individuals interact with their environments over time. By enriching raw trajectories with multi-scale contextual information about location function and neighborhood structure, \systemname constructs semantically meaningful mobility representations that capture both personal behavior patterns and contextual variations. These representations are learned through an LSTM-based clustering architecture that embeds spatial and temporal features into a shared latent space. Building on these representations, \systemname introduces a behavior-cluster-aware modeling mechanism that forms individualized behavioral profiles from normal movement data and detects anomalies through cross-period behavioral comparison with semantic cluster alignment, ensuring consistent interpretation of behavioral shifts across time. 

Our main contributions are threefold: (1) a systematic approach for extracting and encoding multi-scale OSM-based spatial semantics to enhance contextual understanding of mobility behavior; (2) a novel behavior-cluster-aware mechanism for personalized anomaly detection through individualized behavior modeling; and (3) an unsupervised framework that jointly models spatial semantics and temporal dynamics to identify individuals exhibiting meaningful behavioral deviations within large-scale human mobility datasets.

\vspace{-.13in}
\section{Related Work}

Traditionally, unsupervised methods on anomaly detection employ handcrafted features and statistical measures~\cite{zhang2011ibat}. In detail, iBAT~\cite{zhang2011ibat} uses isolation-based analysis, exploiting that anomalous trajectories are ``few and different'', thus easier to isolate. Recent deep learning advances have introduced more sophisticated frameworks for anomaly detection, which can be broadly divided into three strategies. The first is forecasting-based methods like TopoGDN \cite{liu2024multivariate}, which enhanced the graph attention network with multi-scale temporal convolutions and topological feature attention, detecting anomalies through prediction deviations. The second is a reconstruction-based method, such as GM-VSAE~\cite{liu2020online}, which employs Gaussian mixture VAE models for learning trajectory distributions, and context-aware VAE approaches include context-aware anomaly detection with individual-specific embeddings\cite{hu2024context}. In contrast, ATROM~\cite{gao2023open} can be characterized as a contrastive-based approach, as it uses variational Bayesian techniques for behavioral pattern exploration. Beyond these, hybrid methods have also emerged, exemplified by USTAD~\cite{wen2024uncertainty}, which employs dual transformer architectures modeling both aleatoric and epistemic uncertainties in human mobility sequences with uncertainty-aware anomaly scoring. Our method belongs to the hybrid one. Despite these innovations, existing approaches lack systematic multi-scale spatial semantic modeling and individualized behavioral understanding for personalized anomaly detection. We address these gaps through H3-indexed multi-buffer OSM feature extraction and behavior-cluster aware behavioral modeling.

\vspace{-.1in}
\section{Problem Statement}

We define an individual-level mobility anomaly represents a significant deviation from a person's established movement patterns, encompassing behavioral shifts and location-semantic inconsistencies.
Consider an individual $a$ who takes a set of trips, $A = \{T_1, T_2, \ldots, T_N\}$. Each trip $T_j$ consists of a collection of stay points $\mathcal{S} = \{s_{j1}, s_{j2}, \ldots, s_{jM_j}\}$, where each stay point $s_{ji}$ is defined as $s_{ji} = (\text{lat}_{ji}, \text{lon}_{ji}, t^{\text{start}}_{ji}, t^{\text{end}}_{ji}, \text{ID}_a)$. Here, $(\text{lat}_{ji}, \text{lon}_{ji})$ represents the geographic coordinates, $(t^{\text{start}}_{ji}, t^{\text{end}}_{ji})$ denotes the temporal interval, and $\text{ID}_a$ corresponds to the individual $a$ identifier.
We divide the set of trips into two disjoint periods: the past period for training and the future period for testing. Using the past trips, we train a model $f_\theta$ that estimates a distribution of mobility behavior clusters. We then assign each future trip to one of the clusters learned from the past period. To detect anomalies, we compare the alignment of cluster assignments across the two periods. Finally, we compute and output an anomaly score of individual $a$, $\text{anomaly}_a \in [0,1]$.

\begin{figure*}[t]
  \vspace{-.15in}
  \includegraphics[width=0.9\textwidth]{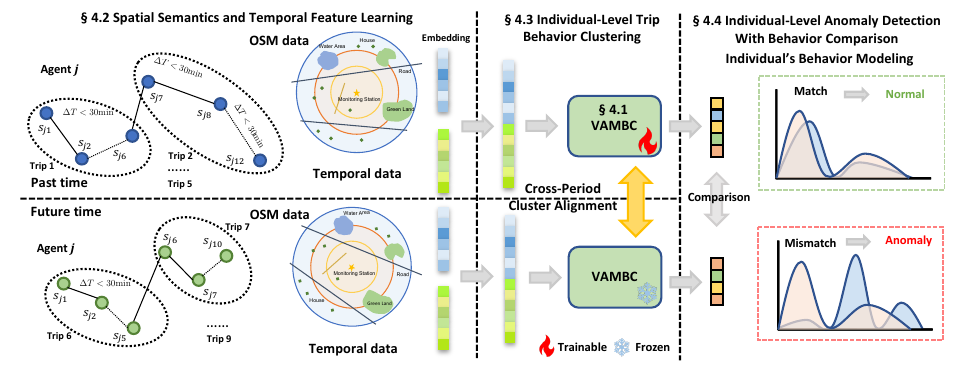}
  \vspace{-.15in} 
  \caption{Overview of \systemname. The framework integrates spatial and temporal features to detect individual-level mobility anomaly detection through cross-period behavioral modeling.
  \vspace{-.1in}
  }
  \label{fig:framework}
\end{figure*}
\vspace{-.1in}

\section{\systemname}
\paragraph{\textbf{Goal}} The goal of \systemname is to detect individual-level behavioral anomalies in human mobility data by learning individualized behavioral patterns from normal movement and identifying deviations through cross-period comparison. By extending existing mobility behavior clustering methods~\cite{yue2021vambc, yue2019detect, hu2022clustering, lin2024unified}, which typically focus on general trajectory-level analysis but have not been effectively adapted for anomaly detection that can capture individual-specific behavioral shifts and location-semantic inconsistencies.

\paragraph{\textbf{Preliminary}} VAMBC~\cite{yue2021vambc}, a variational autoencoder (VAE) architecture-based unsupervised method, for each staypoint, retrieves POIs type frequency of defined buffer as input, produces continuous latent representations (z-mean), and outputs trajectory cluster assignments. VAMBC decomposes the latent representation $z$ into cluster-specific information $z^c$ and individualized bias $z^b$: $z = z^c + z^b$, where $z^c$ captures shared behavioral patterns within clusters. VAMBC optimizes the log-Evidence Lower Bound (ELBO) objective with reconstruction loss to ensure the model can accurately reconstruct input mobility sequences, KL divergence loss to regularize the latent space, and clustering losses including negative entropy loss and center loss to encourage confident cluster assignments and tight intra-cluster cohesion (Equation~\ref{eq:vambc_loss}). 

\small
\begin{equation} \label{eq:vambc_loss}
\mathcal{L} = 
\mathcal{L}_{\text{recon}} + \mathcal{L}_{\text{KL}} + \mathcal{L}_{\text{entropy}} + \mathcal{L}_{\text{center}}
\end{equation}
\normalsize

We develop \systemname upon VAMBC, mobility behavior clustering architecture, and latent representation learning, to learn individualized behavioral modeling and enable precise individual-level anomaly detection.

\vspace{-.1in}
\subsection{Spatial Semantics and Temporal Feature Learning}
\label{sec:spatial-temporal}

Neighboring POIs of staypoints (inputs to VAMBC) may lack rich spatial semantics. To address this, \systemname integrates multi-scale heterogeneous spatial feature learning, capturing point, line, and polygon features from public geographic databases (e.g., OSM), following techniques proposed by Lin et al.~\cite{lin2017mining}.
\systemname constructs multi-scale buffers (e.g., 500 meters, 2000 meters) around each staypoint and computes counts of feature types by category for each buffer. This preserves the functional composition that characterizes different behavioral contexts, enabling detection of semantic anomalies beyond simple location changes. To enable efficient parallel processing, features are indexed using the H3 spatial grid, then projected into a hidden dimension matching the temporal features. Details can be found in ~\cite{lin2017mining}.

For temporal features, the system extracts time-of-day (one-hot), day-of-week (one-hot), time period (one-hot), and continuous features including weekend indicators, duration, and season, along with cyclic encodings (hour and day-of-week sine/cosine transformations). The temporal encoder then applies a linear projection layer to map these temporal features to a unified hidden dimension. Finally, \systemname concatenates both spatial semantics and temporal features of each staypoint along the feature dimension and passes the combined representation through a learnable fusion layer to produce the final unified feature representation for VAMBC processing.

\vspace{-.1in}
\subsection{Individual-Level Trip Behavior Clustering}

The goal is to group trips into clusters that represent distinct lifestyle patterns (e.g., commuting, shopping, recreation). To achieve this, the clustering function $f_\theta$ maps each trip $T_j$ to a cluster ID $c_j \in {1,2,\ldots, K}$, where $K$ is the number of clusters. \systemname uses fused temporal and spatial embeddings of staypoints (Section~\ref{sec:spatial-temporal}) as input and processes each trip through the VAMBC architecture to assign $c_j = f_\theta(T_j)$. By clustering trips across all individuals, the model captures behavioral patterns that generalize across individuals. We extend VAMBC’s reconstruction loss to deal with our spatiotemporal features by separating the loss into temporal and spatial components:

\small
\vspace{-1mm} 
\begin{align}
\mathcal{L}_{\text{recon}} 
&= \alpha \cdot \text{MSE}(x_{\text{temporal}}, \hat{x}_{\text{temporal}}) + \beta \cdot \text{MSE}(x_{\text{spatial}}, \hat{x}_{\text{spatial}})
\end{align}
\vspace{-4mm} 
\normalsize

\systemname trains the clustering function $f_\theta$  only on normal trips from the past period.
For each individual $a$, \systemname constructs behavioral modeling $P_a$ by aggregating their trip-level cluster assignments from these normal periods.
This modeling comprises five key components used for anomaly detection: the cluster distribution $\mathbf{D}_a \in \mathbb{R}^K$ which gives the probability $P(c_k)$ over $K$ mobility clusters; the transition matrix $\mathbf{M}_a \in \mathbb{R}^{K \times K}$ which stores cluster-to-cluster transition probabilities; the dominant cluster $C_a$ which identifies the primary behavioral mode; the entropy $H_a$ which quantifies behavioral complexity; and the trip count which reflects activity frequency. These components encode how individuals typically move and behave, and serve as a reference for detecting anomalies.

\vspace{-.1in}
\subsection{Individual-Level Anomaly Detection with Behavior Clusters}

With the generated mobility behavior clusters of each individual in the past period, here we apply the same trained clustering model to assign cluster labels to trips in the future testing period.
The goal is to, given an individual's test-period trips, construct their test behavioral modeling and compare it against their normal behavioral modeling to detect significant behavioral changes.
If their behavior modelings match, the individual with this given trajectory is considered normal.

A challenge in cross-period analysis is maintaining the semantic consistency of cluster interpretations when cluster IDs learned from training data are applied to test data. We address the challenge through cluster semantic alignment by computing training period cluster centers $\{\mu_1, \mu_2, \ldots, \mu_K\}$ in the latent z-mean space and realigning test period clusters through nearest-neighbor matching:
\small
\vspace{-2mm} 
\begin{align}
c^{aligned} = \arg\min_k \|z - \mu_k\|_2
\end{align}
\normalsize
This ensures that the same cluster IDs represent the same behavioral patterns across time periods, maintaining semantic consistency in cross-temporal analysis. After cross-period cluster alignment, the semantics of the cluster ID of each trip from the training and testing periods are the same. Then, for each individual $a$, we construct test period behaviors and detect individual-level anomalies by comparing them against training behaviors across multiple behavioral dimensions. 

We measure behavioral changes through six dimensions: distribution shifts via Jensen-Shannon divergence $JS(\mathbf{D}_a^{\text{train}}, \mathbf{D}_a^{\text{test}})$, dominant cluster changes $\Delta C_a$ to detect shifts in primary behavioral modes, new behavior emergence $N_a$ to measure the probability mass of previously unseen clusters, changes of the transition pattern $\Delta T_a$ to capture alterations in sequential cluster transitions using the Frobenius norm of transition matrix differences, variations of the entropy $\Delta H_a$ to reflect changes in behavioral complexity and predictability, and frequency changes $\Delta F_a$ to measure relative differences in activity levels between individuals, combining these six behavioral change scores $\text{score}_i$ into a comprehensive weighted sum anomaly score:

\small
\vspace{-4mm} 
\begin{align}
\text{anomaly}_a = \sum_{i} w_i \cdot \text{score}_i ,\tag{4}
\end{align}
\vspace{-4mm} 
\normalsize

where each $\text{score}_i$ represents one of the six behavioral modeling measures and $w_i$ are the corresponding weights.
This approach enables the detection of personalized behavioral anomalies that population-level methods cannot identify, such as significant changes in daily routines or the emergence of entirely new movement patterns, by leveraging individual-specific behavioral modeling established during normal periods.

\vspace{-.1in}
\section{Experiments}

\paragraph{\textbf{Experiment Settings}} We first segment \systemname staypoints into trips using 0.5-hour temporal intervals with a minimum of 4 staypoints per trip. Then, for spatial semantics feature, we use 13 feature types from OSM, including buildings, landuse, transportation infrastructure, water features, points of interest, and natural areas, following previous work~\cite{lin2017mining}. We use three levels of spatial buffer: 500 meters, 1000 meters, and 2000 meters, around each stay point. For spatial indexing, we apply H3 at resolution 10 ($\sim15.3 km^2$ per hexagonal cell). We set the $K$ cluster size of mobility behavior as 6, $\alpha$ as 1.5, and $\beta$ as 1.2. For anomaly score calculation, we use weights $w_1 = 0.25$ for distribution shifts, $w_2 = 0.20$ for new behavior emergence, $w_3 = 0.15$ for transition changes, $w_4 = 0.15$ for entropy variations, $w_5 = 0.15$ for frequency changes, and $w_6 = 0.10$ for dominant cluster changes. These weights are empirically tuned on the NUMOSIM dataset to optimize the model performance.

We evaluate \systemname on NUMOSIM~\cite{stanford2024numosim}, a large-scale synthetic mobility benchmark generated from real travel survey data, which simulates realistic urban movement patterns in Los Angeles for anomaly detection. NUMOSIM provides each individual's trajectories for training and test periods, as well as ground-truth labels of anomalies per individual. We randomly sampled 6,000 individuals from the NUMOSIM dataset while maintaining the original anomaly prevalence rate.
We use the Area Under Receiver Operating Characteristic (AUROC) and Average Precision (AP) as an evaluation metric, following previous work~\cite{wen2024uncertainty, duan2024back}. AUROC measures the separation between normal and anomalous individual behaviors, independent of class imbalance. AP calculates the area under the precision-recall curve, which is especially valuable for evaluating rare event detection in highly imbalanced datasets. We evaluate \systemname against Context-Aware Trajectory Anomaly Detection methods ~\cite{hu2024context} since both use reconstruction-based models and try to include the individual-specific information and spatial context.

\vspace{-.1in}
\paragraph{\textbf{Experiment Results and Discussion}} Table~\ref{tab:baseline_comparison} presents anomaly detection results, showing that Context-Aware Trajectory Anomaly Detection achieves limited performance (AUROC 0.586, AP 0.002), whereas \systemname achieves 0.775 AUROC and 0.0096 AP. 
The high AUROC score is particularly significant for practical anomaly detection applications, as it indicates that \systemname consistently ranks the most unusual behaviors at the top of candidate lists. This ranking capability is often more critical than capturing all anomalies, since practitioners typically focus on investigating the highest-confidence anomalies first.
The relatively low AP primarily stems from the fundamental nature of our behavioral change detection approach. Unlike methods that target only labeled anomalies, \systemname detects shifts in behavioral patterns more broadly. When an individual's routine changes significantly, our method flags it as unusual even if it's not marked as a ground-truth anomaly in the dataset.
\small
\begin{table}[h]
\vspace{-.1in}
\caption{Baseline comparison results using AUROC and AP, where CA-TAD = Context-Aware Trajectory Anomaly Detection~\cite{hu2024context}.}\label{tab:baseline_comparison}
\vspace{-.1in}
\scalebox{1.0}{\begin{tabular}{p{4.5cm}rr}
\toprule
Method & AUROC & AP\\
\midrule
CA-TAD~\cite{hu2024context} & 0.586 & 0.0020\\
\textbf{\systemname (Ours)} & \textbf{0.775} & \textbf{0.0096}\\
\bottomrule
\end{tabular}}
\vspace{-.1in}
\end{table}
\normalsize
We conduct an ablation study to validate each component's contribution (Table~\ref{tab:ablation_results}). The result demonstrates that integrating temporal and spatial semantics enables the detection of anomalies that neither feature type could identify alone.

\small
\begin{table}[h]
\vspace{-.1in}
\caption{Ablation study results on architecture variants using AUROC and AP, where T-only = Temporal-only, S-only = Spatial Semantics-only, and Full = Full Model combining both features.}\label{tab:ablation_results}
\vspace{-.1in}
\scalebox{1.0}{\begin{tabular}{p{1.5cm}ccrr}
\toprule
Architecture & Temporal & Spatial Semantics & AUROC & AP \\
\midrule
T-only & \checkmark & & 0.699 & 0.0085\\
S-only & & \checkmark & 0.757 & 0.0042\\
\textbf{Full} & \checkmark & \checkmark & \textbf{0.775} & \textbf{0.0096}\\
\bottomrule
\end{tabular}}
\vspace{-.1in}
\end{table}
\normalsize

We plan to investigate advanced anomaly scoring techniques to improve precision and reduce false positives in highly imbalanced trajectory datasets. Additionally, we aim to develop more sophisticated cluster alignment mechanisms and automated parameter tuning methods to replace manual weight setting. Furthermore, we plan to evaluate on additional datasets, including real-world mobility datasets and other simulated datasets, to validate the generalizability.

\vspace{-.1in}
\section*{Acknowledgments}
Supported by the Intelligence Advanced Research Projects Activity (IARPA) via the Department of Interior/Interior Business Center (DOI/IBC) contract number 140D0423C0033. The U.S. Government is authorized to reproduce and distribute reprints for Governmental purposes notwithstanding any copyright annotation thereon. Disclaimer: The views and conclusions contained herein are those of the authors and should not be interpreted as necessarily representing the official policies or endorsements, either expressed or implied, of IARPA, DOI/IBC, or the U.S. Government.

\vspace{-.1in}
\bibliographystyle{ACM-Reference-Format}
\bibliography{ref}
\end{document}